\documentclass{article}
\usepackage{iclr2026_conference,times}
\usepackage{amsmath,amsfonts,bm}
\usepackage{hyperref}
\usepackage{url}
\usepackage{graphicx}
\usepackage{booktabs}
\usepackage{xcolor}
\usepackage{soul}
\definecolor{hlred}{RGB}{255,204,204}
\definecolor{hlgreen}{RGB}{204,255,204}
\newcommand{\hlr}[1]{{\sethlcolor{hlred}\hl{#1}}}
\newcommand{\hlg}[1]{{\sethlcolor{hlgreen}\hl{#1}}}
\usepackage{tikz}
\usetikzlibrary{shapes.geometric, arrows.meta, positioning, fit, backgrounds, calc, decorations.pathreplacing}

\newcommand{\method}{Hindsight Preference Optimization}

\iclrfinalcopy

\title{Hindsight Preference Optimization \\ for Financial Time Series Advisory}

\author{%
\textbf{Yanwei Cui\textsuperscript{1}} \quad \textbf{Guanghui Wang\textsuperscript{1}} \quad \textbf{Xing Zhang\textsuperscript{1}} \quad \textbf{Peiyang He\textsuperscript{1}}\thanks{Corresponding author: \texttt{peiyan@amazon.com}} \\
\textbf{Ziyuan Li\textsuperscript{2}} \quad \textbf{Bing Zhu\textsuperscript{2}} \quad \textbf{Wei Qiu\textsuperscript{2}} \quad \textbf{Xusheng Wang\textsuperscript{2}} \quad \textbf{Zheng Yu\textsuperscript{2}} \quad \textbf{Anqi Xin\textsuperscript{2}} \\[0.5em]
\textsuperscript{1}AWS Generative AI Innovation Center \quad  \textsuperscript{2}HSBC Holdings Plc. HSBC Technology Center China
}
\track{Research}

\begin{document}

\maketitle
\thispagestyle{fancy}

\begin{abstract}
Time series models predict numbers; decision-makers need advisory---directional signals with reasoning, actionable suggestions, and risk management. Training language models for such predictive advisory faces a fundamental challenge: quality depends on outcomes unknown at prediction time. We bridge two ideas from reinforcement learning---using information unavailable during execution to retrospectively generate training signal, and preference alignment---and propose \method{}: observed outcomes let an LLM judge rank candidate advisories on dimensions that scalar metrics cannot capture, producing preference pairs for DPO without human annotation. We apply this to Vision-Language-Model-based predictive advisories on S\&P 500 equity time series, demonstrated by a 4B model outperforming its 235B teacher on both accuracy and advisory quality.
\end{abstract}

\section{Introduction}

Recent work has shown that language model architectures transfer remarkably well to time series forecasting \citep{ansari2024chronos, das2024decoder, shi2026kronos}, positioning pretrained models as powerful tools for temporal prediction. Yet these models output tokenized numbers, not natural language. Decision-makers need more: not just ``the price will be \$152'' but why (reasoning about patterns), how confident (calibrated uncertainty), and what to do (actionable suggestions with risk management). This structured advisory requires natural language generation, motivating language models that can process historical data and produce reasoned, actionable output.

Vision-language models (VLMs) offer a natural bridge: they can interpret financial charts and generate textual analysis \citep{bai2025qwen3vl}. However, training VLMs for predictive advisory introduces a challenge distinct from typical vision-language tasks. Unlike image captioning or visual question answering, where quality can be judged immediately, advisory quality depends on outcomes that unfold after prediction.

Our approach bridges two ideas from reinforcement learning: the retrospective training signal---using information unavailable during execution to generate supervision after the fact \citep{andrychowicz2017her}---and preference alignment, where RLHF and its scalable variants---RLAIF \citep{lee2024rlaif}, LLM-as-a-Judge \citep{zheng2023judging}---align models using preference pairs. We extend this framework by providing the judge with ground-truth outcomes, enabling objective evaluation of predictive advisory. The resulting framework, \method{}, automatically constructs preference pairs for DPO \citep{rafailov2023dpo} from observed outcomes---eliminating human annotation while rewarding analytical quality over coincidental correctness.

Our contributions: (1) \textbf{\method{}}---grounding LLM-as-a-Judge evaluation in observed outcomes to assess structured predictions---scenarios, risk estimates, and reasoning---beyond what scalar metrics can capture; and (2) \textbf{VLM-based financial advisory}---applying our framework to train a 4B model that interprets candlestick charts and generates reasoned advisory, outperforming its 235B teacher on both accuracy and advisory quality.

\begin{figure}[t]
\centering
\begin{tikzpicture}[
    box/.style={rectangle, thick, minimum height=1.3cm, minimum width=2.8cm, align=center, font=\small, rounded corners=2pt},
    inputbox/.style={box, draw=black!50, fill=gray!5},
    advbox/.style={box, draw=blue!60, fill=blue!8, minimum width=3.4cm},
    outbox/.style={box, draw=green!60!black, fill=green!8},
    judgebox/.style={box, draw=orange!70!black, fill=orange!8, minimum width=5cm},
    arr/.style={-{Stealth[length=2mm]}, thick, black!50},
    timeline/.style={very thick, black!30},
    tlabel/.style={font=\small\bfseries, black!70},
    sublabel/.style={font=\scriptsize, black!50},
]

\draw[timeline] (-5.5, 0) -- (5.5, 0);
\node[tlabel, above=2pt] at (-4, 0) {Context Window};
\node[tlabel, above=2pt] at (0, 0) {Prediction Time};
\node[tlabel, above=2pt] at (4, 0) {Outcome Window};
\fill[black!40] (-4, 0) circle (3pt);
\fill[blue!60] (0, 0) circle (3pt);
\fill[green!50!black] (4, 0) circle (3pt);

\node[sublabel] at (0, -0.65) {$t$};

\draw[decorate, decoration={brace, amplitude=5pt, mirror}, black!40] (-5.3, -0.25) -- (-2.7, -0.25);
\node[sublabel] at (-4, -0.65) {historical data $t{-}T$};
\draw[decorate, decoration={brace, amplitude=5pt, mirror}, black!40] (2.7, -0.25) -- (5.3, -0.25);
\node[sublabel] at (4, -0.65) {prediction horizon $t{+}\Delta$};

\node[inputbox] (chart) at (-4, -1.8) {\textbf{Input} $x$\\{\scriptsize candles chart}};
\node[advbox] (adv) at (0, -1.8) {\textbf{Candidates} $\{y_1, \ldots, y_K\}$\\{\scriptsize advisories: actionable scenarios,}\\{\scriptsize reasoning, risk estimates}};
\node[outbox] (out) at (4, -1.8) {\textbf{Outcome} $o$\\{\scriptsize actual price movement}};

\draw[arr] (chart.east) -- (adv.west);

\node[judgebox] (judge) at (0, -3.6) {\textbf{LLM Judge}: rank $y_{(1)} \succ \cdots \succ y_{(K)}$\\{\scriptsize compare predictions against actual $o$}};

\draw[arr] (adv.south) -- (judge.north);
\draw[arr, green!60!black] (out.south) |- node[pos=0.25, right, sublabel, green!50!black] {hindsight} (judge.east);

\node[font=\small\bfseries, align=center] (output) at (0, -5.2) {$\bm{y^+ = y_{(1)}, \; y^- = y_{(K)}}$\\{\small Hindsight Preference Generation}};
\draw[arr, orange!60!black] (judge.south) -- (output.north);

\end{tikzpicture}
\caption{\method{} framework. The LLM judge ranks candidate advisories against observed outcomes, producing preference pairs for DPO.}
\label{fig:framework}
\end{figure}
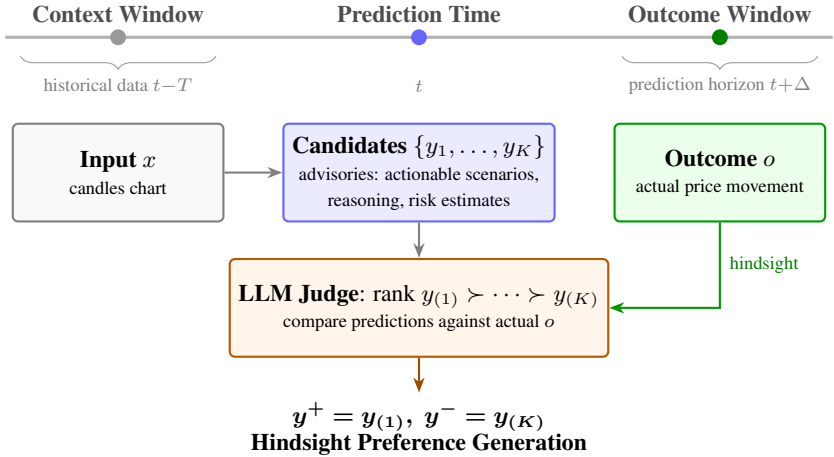

\section{Related Work}

\textbf{Time Series and Language Models.} Recent work reveals that language model architectures generalize effectively to temporal data. Chronos \citep{ansari2024chronos} and TimesFM \citep{das2024decoder} demonstrate that tokenizing time series values enables transformers to achieve strong zero-shot forecasting. More relevant to our work, Kronos \citep{shi2026kronos} targets financial markets specifically, pre-training on billions of candlestick records to learn price dynamics and trading patterns---though it outputs numerical predictions rather than natural language advisory. Vision-language models extend forecasting to visual inputs: Qwen3-VL \citep{bai2025qwen3vl} generates textual analysis from images. In finance, domain-specific LLMs have shown promise for market text analysis \citep{li2023large}. We build on these advances, combining chart understanding with structured advisory generation.

\textbf{Preference Learning with Hindsight.} Direct Preference Optimization \citep{rafailov2023dpo} enables training from preference pairs without explicit reward modeling. RLAIF \citep{lee2024rlaif} and LLM-as-a-Judge \citep{zheng2023judging} scale preference collection through AI feedback, proving effective for subjective qualities. The idea of retrospective training signal---using post-hoc knowledge to generate supervision unavailable during execution---originates in reinforcement learning \citep{andrychowicz2017her}. Recent work such as mDPO \citep{wang2024mdpo} and RPO \citep{zhao2025reflective} confirms that enriching DPO pairs with external signal improves alignment in vision-language models for reducing hallucination. We bring this principle to preference learning, grounding evaluation in observed outcomes for predictive tasks.

\section{Method}

\subsection{Problem Setup}

Let $x \in \mathcal{X}$ denote the input context, which may include multiple modalities (e.g., time series data, charts, text). Let $y \in \mathcal{Y}$ denote the generated advisory containing predictive claims---directional forecasts, confidence levels, specific targets---along with reasoning and actionable guidance.

We also have $o \in \mathcal{O}$, the ground-truth outcome, which is unavailable at prediction time but can be collected after the prediction window. Unlike standard forecasting that minimizes prediction error against $o$, we use $o$ to enable comparative evaluation: the judge ranks advisories by how well they anticipated the outcome, producing preference pairs for DPO rather than scalar losses.

\subsection{Hindsight Preference Generation}

Given candidates $\{y_1, \ldots, y_K\}$ generated from input $x$ (Figure~\ref{fig:framework}), an LLM-as-a-Judge $J$ evaluates each against the ground-truth outcome $o$, producing a ranking $y_{(1)} \succ \cdots \succ y_{(K)}$. Each advisory contains freeform reasoning alongside structured predictions: multiple scenarios, each describing a predicted price path (e.g., steady decline, V-shaped recovery) with an associated likelihood, as well as quantitative risk estimates such as expected drawdown and volatility. These are not assumed to be calibrated---they are the model's expressed assessments, which the judge evaluates against realized outcomes. Traditional scalar metrics such as MAE or MSE can evaluate point predictions but cannot assess reasoning quality, scenario calibration, or risk management; an LLM-as-a-Judge operating with hindsight can evaluate these richer dimensions jointly.

With access to outcomes unavailable at prediction time, the judge can distinguish sound analysis from lucky guesses---the core advantage of hindsight. It jointly evaluates \emph{directional signals} (scenario-outcome alignment), \emph{reasoning} (whether identified dynamics actually drove the move), and \emph{risk management} (drawdown and volatility calibration). The resulting preference pairs encode not just \emph{what} to predict but \emph{how} to reason, derived automatically without human annotation.

\subsection{Hindsight-Guided Preference Optimization}

\textbf{Stage 1: Supervised Fine-Tuning.} We adopt vision-language models (VLMs) as the base architecture, leveraging their pretrained capabilities for chart understanding and textual reasoning. A large teacher model generates $K$ candidate advisories for each input $x$. Using hindsight ranking, the judge selects top-ranked candidates as supervision targets $y^*$. A smaller student model is trained to reproduce these hindsight-selected examples:
\begin{equation}
    \mathcal{L}_{\text{SFT}}(\theta) = -\mathbb{E}_{(x, y^*)} \left[ \log \pi_\theta(y^* \mid x) \right]
    \label{eq:sft}
\end{equation}
This bootstraps the student from the teacher's best candidates.

\textbf{Stage 2: Hindsight DPO.} We refine the student model through on-policy preference optimization: (1) sample $K$ candidates from the student model, (2) rank against outcomes using the judge, (3) construct preference pairs with $y^+ = y_{(1)}$ and $y^- = y_{(K)}$, and (4) optimize using DPO. This process naturally extends to multiple rounds of iterative refinement:
\begin{equation}
    \mathcal{L}_{\text{DPO}}(\theta) = -\mathbb{E}_{(x, y^+, y^-)} \left[ \log \sigma \left( \beta \left( \log \frac{\pi_\theta(y^+ \mid x)}{\pi_{\text{ref}}(y^+ \mid x)} - \log \frac{\pi_\theta(y^- \mid x)}{\pi_{\text{ref}}(y^- \mid x)} \right) \right) \right]
    \label{eq:dpo}
\end{equation}
We augment with anchored regularization on chosen responses \citep{wang2024mdpo, zhao2025reflective}, following recent practices in preference optimization:
\begin{equation}
    \mathcal{L}(\theta) = \mathcal{L}_{\text{DPO}}(\theta) + \alpha \cdot \mathcal{L}_{\text{SFT}}(y^+ \mid x)
    \label{eq:total}
\end{equation}
The DPO term isolates what distinguishes $y^+$ from $y^-$---shared surface-level variations cancel out in the log-likelihood ratio, leaving the patterns that drive preference. The SFT anchor on $y^+$ prevents likelihood degradation as preferences shift the distribution.

\section{Experiments and Discussion}

\textbf{Setup.} We evaluate our framework on S\&P 500 equities using daily OHLC (Open-High-Low-Close) data spanning 2013--2017 \citep{nugent2018sp500}. Each sample pairs a candlestick chart showing one month of historical trading with the ground-truth price movements from the subsequent week. Charts contain only price candles and volume bars with no ticker symbols, dates, or axis labels that would identify the security or time period. The model generates structured advisory for this future window: directional scenarios with reasoning and risk estimates. We select the top 5 S\&P 500 stocks and use a temporal split: data from 2013--2016 serves as the training set (1{,}395 samples at 5-day intervals), while all of 2017 is held out for testing (365 samples at 5-day intervals).
We use Qwen3-VL-235B as the teacher model for generating candidate advisories during SFT data collection, and Qwen3-VL-4B as the student model that undergoes training. For preference pair construction (both SFT selection and DPO ranking), the LLM judge is Pixtral Large 25.02; for pairwise preference evaluation (Table~\ref{tab:preference}), we use Claude Opus 4. Training uses LoRA on 4$\times$ L40S GPUs (g6e.12xlarge) with batch size 16 for both SFT and DPO.

We assess performance using three metrics. \emph{Directional Accuracy}: percentage of correct directional predictions, excluding moves within $\pm$1\% as ambiguous (235 of 365 test samples). \emph{Top-1 Scenario Accuracy}: whether the model's top predicted scenario (e.g., \texttt{RALLY\_THEN\_FADE}, \texttt{V\_RECOVERY}, among 5 labels) matches the realized 5-day price path; computed on all 365 test samples. \emph{Pairwise Preference}: an LLM judge with access to outcomes ranks advisory from two models by actionable guidance, reasoning, and risk management. We report means over 5 evaluation runs.

\begin{table}[t]
\caption{Prediction quality on held-out 2017 data (mean $\pm$ std, 5 evaluation runs).}
\label{tab:accuracy}
\centering
\small
\begin{tabular}{@{}lcc@{}}
\toprule
\textbf{Model} & \textbf{Directional Acc.} & \textbf{Top-1 Scenario Acc.} \\
\midrule
Claude Sonnet 3.7 (zero-shot) & 56.1\% $\pm$ 0.8\% & 27.3\% $\pm$ 2.4\% \\
Qwen3-VL-235B (zero-shot) & 51.9\% $\pm$ 1.4\% & 23.7\% $\pm$ 1.5\% \\
Qwen3-VL-4B (zero-shot) & 50.7\% $\pm$ 1.8\% & 22.1\% $\pm$ 1.3\% \\
\midrule
Qwen3-VL-4B + SFT (Stage 1) & 53.3\% $\pm$ 2.7\% & 25.2\% $\pm$ 1.8\% \\
Qwen3-VL-4B + Hindsight DPO & 57.9\% $\pm$ 2.3\% & 27.1\% $\pm$ 2.0\% \\
\bottomrule
\end{tabular}
\end{table}

\begin{table}[t]
\caption{Pairwise preference win rates on held-out 2017 data (mean $\pm$ std, 5 evaluation runs).}
\label{tab:preference}
\centering
\small
\begin{tabular}{@{}lccc@{}}
\toprule
& \multicolumn{3}{c}{\textbf{Win Rate Against}} \\
\cmidrule(l){2-4}
\textbf{Model} & \textbf{Qwen3-VL-235B} & \textbf{Claude Sonnet 3.7} & \textbf{Qwen3-VL-4B} \\
\midrule
Qwen3-VL-4B + SFT (Stage 1) & 48.7\% $\pm$ 1.3\% & 46.7\% $\pm$ 1.4\% & 62.5\% $\pm$ 1.8\% \\
Qwen3-VL-4B + Hindsight DPO & 56.8\% $\pm$ 2.7\% & 52.6\% $\pm$ 0.8\% & 71.4\% $\pm$ 2.1\% \\
\bottomrule
\end{tabular}
\end{table}

\textbf{Results.} Table~\ref{tab:accuracy} and Table~\ref{tab:preference} present our main results. Per-class precision/recall, dataset distribution, example model inputs, and a detailed advisory comparison are provided in Appendices~\ref{app:dataset}--\ref{app:advisory}.

SFT on hindsight-selected examples teaches the 4B student structured reasoning and advisory format, achieving 53.3\% directional accuracy and 25.2\% scenario accuracy (Table~\ref{tab:accuracy})---both comparable to the 235B teacher (51.9\% and 23.7\%; McNemar's test, $p{>}0.05$). Pairwise preference confirms parity: the SFT model achieves only 48.7\% win rate against the teacher (Table~\ref{tab:preference}). Even boosted with top-ranked candidates, SFT distills the teacher's capabilities but cannot surpass it---the contrastive signal needed to distinguish better reasoning is absent.

Hindsight DPO bridges this gap. By contrasting advisories ranked against observed outcomes, DPO shifts the model toward patterns the judge associates with better predictive advisory. After preference optimization, the model achieves 57.9\% directional accuracy and 27.1\% scenario accuracy, significantly outperforming both SFT and the 235B teacher on both metrics (McNemar's test, $p{<}0.05$). Pairwise win rates also improve to 56.8\% against Qwen3-VL-235B. As reference points, the improvement over the 4B zero-shot base is decisive (71.4\% win rate), and the DPO-trained model achieves a 52.6\% win rate against Claude Sonnet 3.7 (sign test, $p{<}0.05$).

\textbf{Limitations.} Our evaluation is limited to S\&P 500 equities using only visual price and volume patterns; market outcomes also depend on news, earnings, and macroeconomic factors not captured in charts. The 2013--2017 evaluation window is predominantly bullish, and performance in volatile or bearish regimes (e.g., 2008, 2020, 2022) remains untested. Our goal is not to outperform the market but to improve the quality of structured advisory---reasoning, scenario calibration, and risk awareness---for decision support. Furthermore, both preference construction and evaluation rely on LLM judges without human expert validation; real-world financial adoption would require domain expert assessment to ensure advisory quality aligns with practitioner standards. Additionally, the current approach uses only the judge's rankings, which provides a weak learning signal for DPO---leading to slow and unstable convergence. Incorporating judge rationales as natural-language reward signals and disentangling which advisory dimensions---scenario calibration, reasoning quality, or risk estimation---drive improvement are promising directions.

\section{Conclusion}

Hindsight Preference Optimization bridges two ideas from reinforcement learning---retrospective training signal and preference alignment---for predictive advisory. Armed with observed outcomes, an LLM-as-a-Judge evaluates structured advisory along dimensions inaccessible to scalar metrics: directional signals, reasoning, actionable suggestions, and risk management. The resulting preference pairs for DPO require no human annotation. Our experiments show that prediction accuracy and advisory quality can be jointly optimized, enabling a 4B model to outperform its 235B teacher on both dimensions. The framework applies broadly: any predictive domain where outcomes are eventually observed can benefit from hindsight-grounded preference learning.

\bibliographystyle{iclr2026_conference}
\bibliography{references}

\appendix

\section{Dataset Details}
\label{app:dataset}

Table~\ref{tab:data_dist} summarizes the held-out 2017 evaluation set. The 365 samples across 5 tickers (AAPL, AMZN, FB, GOOGL, MSFT; 73 per ticker) exhibit a bullish skew reflective of the 2017 market: 43.8\% of 5-day windows are classified as Bullish (${\geq}$1\% gain), 20.5\% as Bearish (${\leq}$-1\% loss), and 35.6\% as Neutral. After excluding the 130 neutral samples, the 235 directional samples used for accuracy evaluation have a 68.1\%/31.9\% Bullish/Bearish class imbalance.

\begin{table}[h]
\caption{Evaluation set distribution (held-out 2017, 365 samples across 5 tickers: AAPL, AMZN, FB, GOOGL, MSFT).}
\label{tab:data_dist}
\centering
\small
\begin{tabular}{@{}llcc@{}}
\toprule
\textbf{Direction} & \textbf{Criterion} & \textbf{Count} & \textbf{\%} \\
\midrule
Bullish & ${\geq}$ +1\% & 160 & 43.8\% \\
Bearish & ${\leq}$ $-$1\% & 75 & 20.5\% \\
Neutral (excluded) & within $\pm$1\% & 130 & 35.6\% \\
\midrule
\textbf{Directional (evaluated)} & & \textbf{235} & \textbf{64.4\%} \\
\bottomrule
\end{tabular}
\end{table}

Table~\ref{tab:confusion} reports per-class precision and recall on the 235 directional samples.

\begin{table}[h]
\caption{Per-class precision and recall on held-out 2017 directional samples (mean $\pm$ std, 5 runs).}
\label{tab:confusion}
\centering
\small
\begin{tabular}{@{}lcccc@{}}
\toprule
& \multicolumn{2}{c}{\textbf{Bullish}} & \multicolumn{2}{c}{\textbf{Bearish}} \\
\cmidrule(lr){2-3} \cmidrule(lr){4-5}
\textbf{Model} & \textbf{Prec.} & \textbf{Recall} & \textbf{Prec.} & \textbf{Recall} \\
\midrule
Claude Sonnet 3.7 (zero-shot) & 66.5 $\pm$ 0.7 & 71.5 $\pm$ 1.0 & 27.6 $\pm$ 2.2 & 23.2 $\pm$ 2.8 \\
Qwen3-VL-235B (zero-shot) & 66.2 $\pm$ 1.2 & 60.0 $\pm$ 1.2 & 28.9 $\pm$ 2.1 & 34.7 $\pm$ 3.3 \\
\midrule
Qwen3-VL-4B + SFT (Stage 1) & 66.3 $\pm$ 1.8 & 63.8 $\pm$ 3.4 & 28.6 $\pm$ 3.4 & 30.9 $\pm$ 4.0 \\
Qwen3-VL-4B + Hindsight DPO & \textbf{68.0} $\pm$ 1.5 & \textbf{72.0} $\pm$ 3.0 & \textbf{31.7} $\pm$ 3.9 & 27.7 $\pm$ 4.8 \\
\bottomrule
\end{tabular}
\end{table}

\section{Example Input}
\label{app:input}

\begin{figure}[h]
\centering
\includegraphics[width=0.85\textwidth]{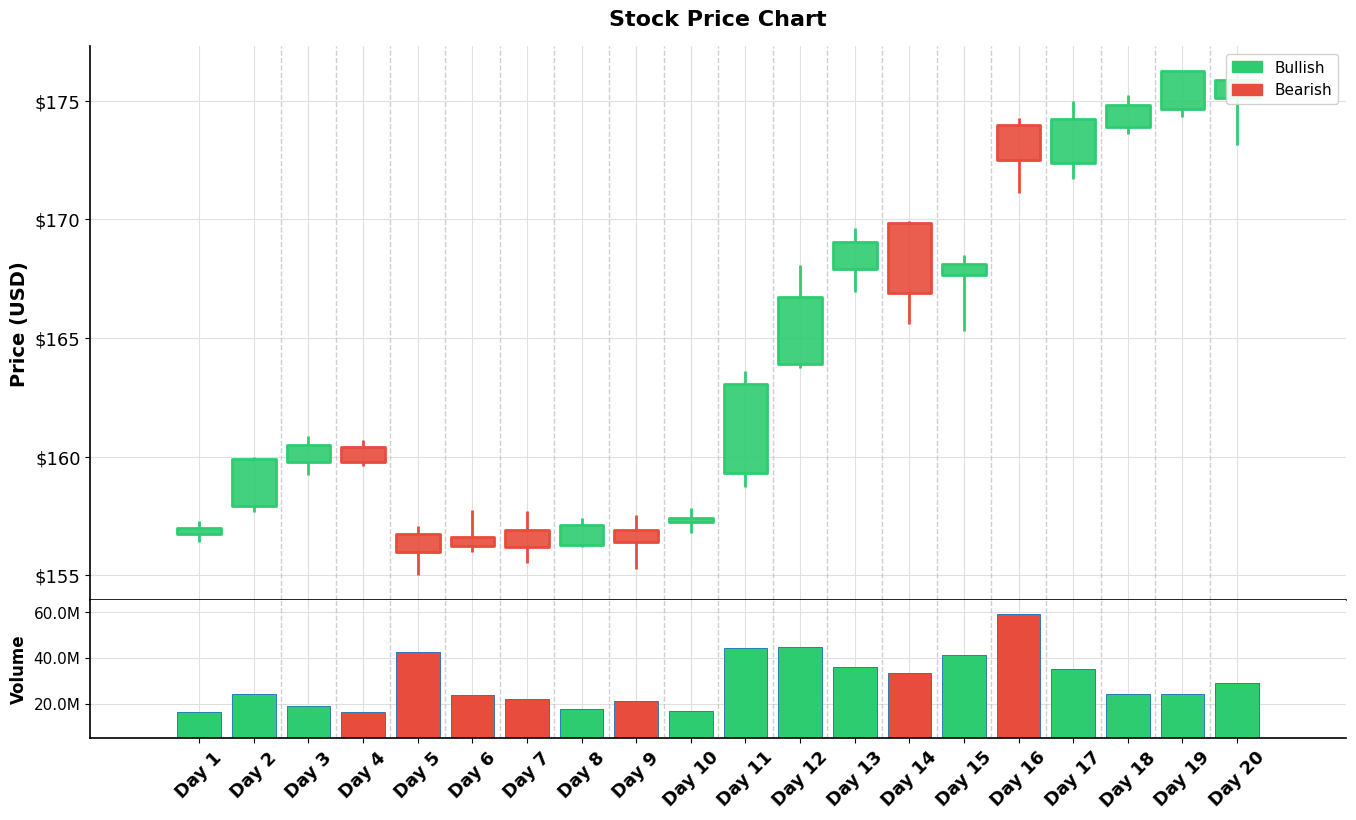}
\caption{Model input: 20-day candlestick chart with price and volume. The model generates advisory from this chart alone.}
\label{fig:chart_input}
\end{figure}

Figure~\ref{fig:chart_input} shows the candlestick chart provided to the VLM at inference time (20 trading days of historical data). The model generates structured advisory based solely on this visual input---no ticker symbols, dates, or axis labels that would identify the security or time period are provided.

\begin{figure}[h]
\centering
\includegraphics[width=0.85\textwidth]{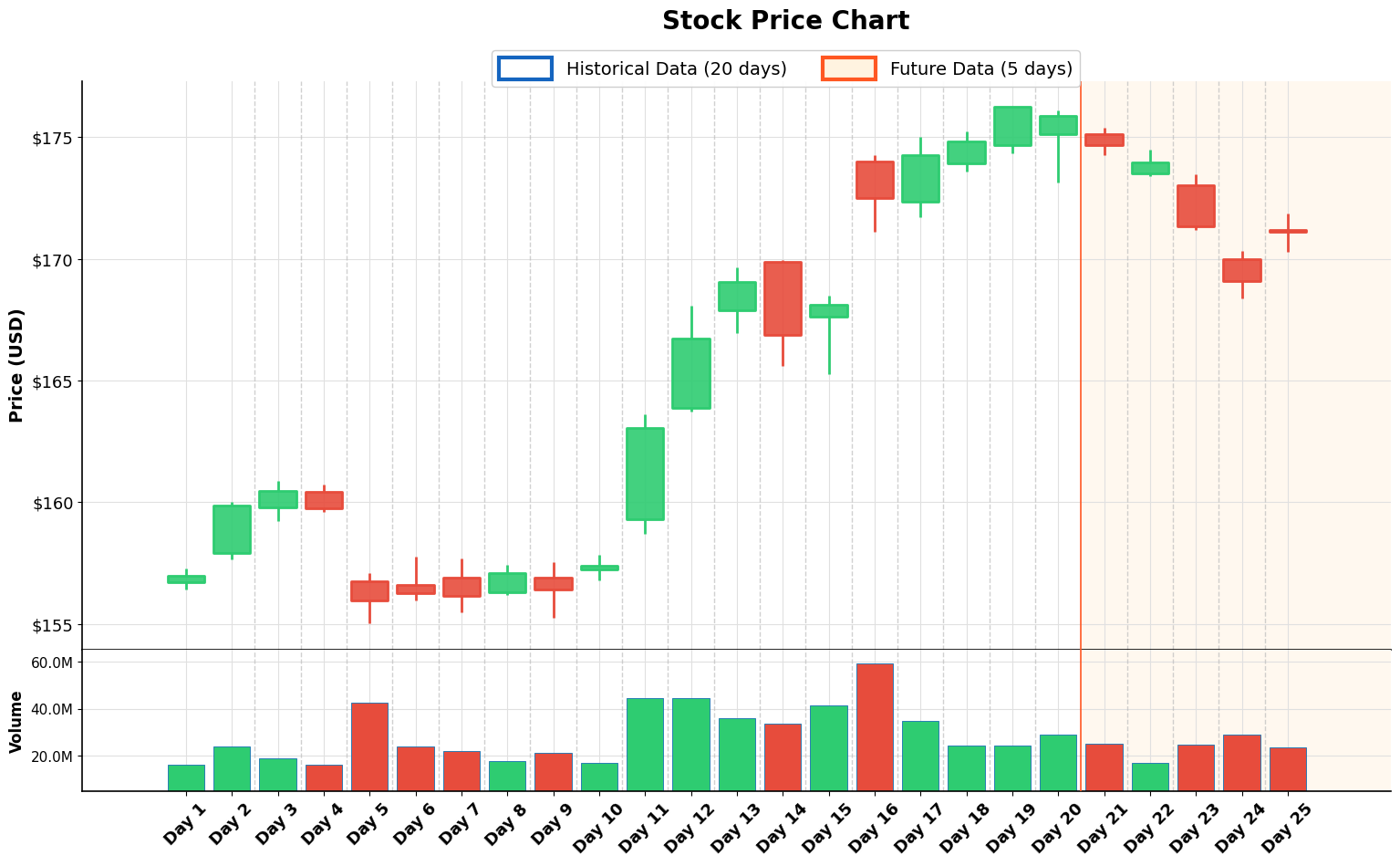}
\caption{Judge input: the same chart extended with the 5-day outcome window (shaded region, Days 21--25). The judge uses this hindsight to rank candidate advisories.}
\label{fig:chart_full}
\end{figure}

Figure~\ref{fig:chart_full} shows the extended chart used by the LLM judge during training. The shaded region (Days 21--25) reveals the 5-day outcome, enabling the judge to evaluate advisory quality with hindsight and construct preference pairs for DPO.

\section{Advisory Comparison}
\label{app:advisory}

We compare advisories generated by three models on the same input (the candlestick chart in Figure~\ref{fig:chart_input}). The ground truth outcome was \textbf{BEARISH} ($-2.72\%$, max drawdown $-4.26\%$). Table~\ref{tab:advisory_comparison} summarizes the predictions.

\begin{table}[h]
\caption{Advisory comparison on a held-out AAPL sample. Ground truth: BEARISH ($-2.72\%$, max drawdown $-4.26\%$).}
\label{tab:advisory_comparison}
\centering
\small
\begin{tabular}{@{}lccc@{}}
\toprule
& \textbf{Qwen3-VL-235B} & \textbf{Claude Sonnet 3.7} & \textbf{Hindsight DPO (4B)} \\
\midrule
Outlook & BULLISH & BULLISH & \textbf{BEARISH} \checkmark \\
Top scenario & STEADY\_UP (55\%) & RALLY\_THEN\_FADE (40\%) & RALLY\_THEN\_FADE (45\%) \\
Selloff probability & 5\% & 10\% & 35\% \\
Max drawdown est. & $-2.5\%$ & $-3.5\%$ & $-2.5\%$ \\
\bottomrule
\end{tabular}
\end{table}

Below we show the full advisory output from each model, including directional signals, reasoning, actionable suggestions, and risk management. \hlr{Red highlights} mark incorrect or overconfident bullish claims; \hlg{green highlights} mark correct bearish signals aligned with the outcome.

\smallskip
\noindent\textbf{Qwen3-VL-235B (zero-shot) --- Outlook: BULLISH}
\begin{quote}
\small
The 20-day candlestick chart displays a clear uptrend interrupted by two key periods of consolidation and minor reversal. [\ldots] A decisive breakout occurs on Day 11, with a large green candle and strong volume, followed by a healthy pullback on Day 14 --- signaling potential support. The subsequent rebound on Days 15--20 shows resilience: higher highs, higher lows, and steady volume. [\ldots] \hlr{The bullish interpretation is stronger: price consistently made higher highs after the Day 11 breakout,} \hlr{volume supported rallies, and corrections were shallow and short-lived.} \hlr{The market structure is not showing signs of topping --- no major distribution or bearish divergence.} [\ldots] STEADY\_UP (55\%), SELLOFF\_THEN\_STABILIZE (5\%).

\smallskip
\noindent\textit{Structured advisory:} Scenarios: \hlr{STEADY\_UP (55\%)}, RALLY\_THEN\_FADE (25\%), SIDEWAYS (15\%), \hlg{SELLOFF\_THEN\_STABILIZE (5\%)}. Confidence: 0.70. Volatility: MODERATE. Key trigger: Price holds above \$170 on Day 21. Risk factor: Breakdown below \$170 with increasing volume. Max drawdown: $-2.5\%$.
\end{quote}

\smallskip
\noindent\textbf{Claude Sonnet 3.7 (zero-shot) --- Outlook: BULLISH}
\begin{quote}
\small
The stock has demonstrated a strong overall uptrend, moving from approximately \$157 to \$176, representing about a 12\% gain over the 20-day period. [\ldots] \hlg{Recent bullish days (17--20) show moderately strong but decreasing volume, suggesting potential} \hlg{momentum fatigue} --- \hlr{yet the conclusion remains: ``a buy-the-dip mentality among market participants.''} [\ldots] The pullback on Day 14 and Day 16 came with elevated volume, but bulls quickly reasserted control. [\ldots] STEADY\_UP (30\%), SELLOFF\_THEN\_STABILIZE (10\%).

\smallskip
\noindent\textit{Structured advisory:} Scenarios: RALLY\_THEN\_FADE (40\%), \hlr{STEADY\_UP (30\%)}, SIDEWAYS (20\%), \hlg{SELLOFF\_THEN\_STABILIZE (10\%)}. Confidence: 0.75. Volatility: MODERATE. Key trigger: Price action around \$178--180 resistance. Risk factor: Exhaustion of buying momentum after extended rally. Max drawdown: $-3.5\%$.
\end{quote}

\smallskip
\noindent\textbf{Hindsight DPO (4B) --- Outlook: BEARISH \checkmark}
\begin{quote}
\small
The 20-day candlestick chart reveals a stock that has experienced a strong uptrend from approximately \$155 to \$175. [\ldots] However, the most recent price action (Days 17--20) is concerning: \hlg{despite closing higher on Day 19 and 20, the candles are becoming increasingly smaller, and the volume} \hlg{is not expanding on up days --- a potential sign of weakening conviction.} \hlg{Day 16 is a large bearish candle that pulled back sharply after a rally, followed by a series of small-bodied candles with diminishing volume on Days 18--20. This suggests exhaustion or distribution at current levels.} [\ldots] \hlg{The failure to break above \$175 with conviction (no follow-through volume) may signal a} RALLY\_THEN\_FADE. [\ldots] RALLY\_THEN\_FADE (45\%), SELLOFF\_THEN\_STABILIZE (35\%).

\smallskip
\noindent\textit{Structured advisory:} Scenarios: RALLY\_THEN\_FADE (45\%), \hlg{SELLOFF\_THEN\_STABILIZE (35\%)}, SIDEWAYS (20\%). Confidence: 0.65. Volatility: MODERATE. Key trigger: Stock opens above \$175 on Day 1 with increasing volume. Risk factor: Positive catalyst reignites buying interest. Max drawdown: $-2.5\%$.
\end{quote}

\end{document}